\documentclass[lettersize,journal]{IEEEtran}
\usepackage{amsmath,amsfonts}
\usepackage{algorithmic}
\usepackage{algorithm}
\usepackage{array}
\usepackage[caption=false,font=normalsize,labelfont=sf,textfont=sf]{subfig}
\usepackage{textcomp}
\usepackage{stfloats}
\usepackage{url}
\usepackage{verbatim}
\usepackage{graphicx}
\usepackage{cite}
\usepackage{xspace}
\usepackage{multirow}

\usepackage{booktabs}
\usepackage{amssymb}

\usepackage{bm}

\begin{document}

\title{Attention in Attention: Modeling Context Correlation for Efficient Video Classification}

\author{
Yanbin Hao,~\IEEEmembership{Member,~IEEE}, 
Shuo Wang, 
Pei Cao, 
Xinjian Gao,~\IEEEmembership{Member,~IEEE},  
Tong Xu,~\IEEEmembership{Member,~IEEE},  
Jinmeng Wu, 
Xiangnan He,~\IEEEmembership{Member,~IEEE}
\thanks{Y. Hao, S. Wang(corresponding author) and X. He are with the CCCD Key Lab of Ministry of Culture and Tourism, School of Data Science, School of Information Science and Technology, University of Science and Technology of China, Anhui, 230026, China.  E-mail: haoyanbin@hotmail.com, shuowang.hfut@gmail.com, xiangnanhe@gmail.com.}
\thanks{P. Cao is with the Wuhan Research Institute of Posts and Telecommunications, Wuhan, Hubei, 430205, China. E-mail: meii11cao@gmail.com.} 
\thanks{X. Gao is with the School of Computer Science and Information Engineering, School of Artificial Intelligence, Hefei University of Technology, Anhui, 230009, China. E-mail: gao\_xinjian@outlook.com. }
\thanks{T. Xu is with the School of Data Science, School of Computer Science and Technology, University of Science and Technology of China, Anhui, 230026, China.  E-mail: tongxu@ustc.edu.cn.}
\thanks{J. Wu is with the Hubei Key Laboratory of Optical Information and Pattern Recognition, Wuhan Institute of Technology Wuhan, Hubei, 430070, China. E-mail: jinmeng@wit.edu.cn.}
}



\maketitle

\begin{abstract}
Attention mechanisms have significantly boosted the performance of video classification neural networks thanks to the utilization of perspective contexts. However, the current research on video attention generally focuses on adopting a specific aspect of contexts (e.g., channel, spatial/temporal, or global context) to refine the features and neglects their underlying correlation when computing attentions. This leads to incomplete context utilization and hence bears the weakness of limited performance improvement. To tackle the problem, this paper proposes an efficient attention-in-attention (AIA) method for element-wise feature refinement, which investigates the feasibility of inserting the channel context into the spatio-temporal attention learning module, referred to as CinST, and also its reverse variant, referred to as STinC. Specifically, we instantiate the video feature contexts as dynamics aggregated along a specific axis with global average and max pooling operations. The workflow of an AIA module is that the first attention block uses one kind of context information to guide the gating weights calculation of the second attention that targets at the other context. Moreover, all the computational operations in attention units act on the pooled dimension, which results in quite few computational cost increase ($<$0.02\%). To verify our method, we densely integrate it into two classical video network backbones and conduct extensive experiments on several standard video classification benchmarks. The source code of our AIA is available at \url{https://github.com/haoyanbin918/Attention-in-Attention}.
\end{abstract}


\section{Introduction}
\label{intro}
Convolutional neural networks (CNNs) have become the de-facto standard for visual content understanding in computer vision communities \cite{he2016deep,wang2016hierarchical,xu2018dual}. Currently, along with the rising demand for video data processing, video CNNs (also called 3D CNNs) have stepped into a prosperous age in the past few years. Most standard 2D and 3D CNNs achieve entire visual content reception through replicating stylized spatial/spatio-temporal convolutions. Although sufficient replicates theoretically enlarge the receptive field to cover the image/video in full, the effective size of such fields is, in practice, considerably smaller \cite{hu2018gather}, especially in the previous layers. This has been repeatedly corroborated by the significant performance gains of incorporating global contexts to feature learning (aka feature contextualization) in various image and video understanding tasks \cite{hu2018squeeze,hu2018gather,xie2018rethinking,li2020tea,liu2020tam}. This is because that visual content can be flexibly analyzed from different perspectives or axes, e.g., spatial, temporal and spatio-temporal, and thereby giving attention to the axial context can enhance the plain feature with such as long-range dependencies.

Current attention-based feature contextualization approaches mainly focus on adjusting the plain features (4-dimensional $T\times H\times W\times C$ tensors) obtained from CNN models (e.g., C3D \cite{tran2015learning}, I3D \cite{carreira2017quo}, CNN-RNN \cite{li2021spatio}, coupled-CNN \cite{wu2018learning}) with different types of spatio-temporal contexts. Generally, the attentional contexts could be information dynamics aggregated from axes (e.g. $T, H, W$) and are for $C$ channel. For instance, squeeze-and-excitation network (SE-Net) \cite{hu2018squeeze} is the first one to use the dynamics aggregated along space axes to contextualize the feature channels. Separable 3D convolution with gating (S3D-G) \cite{xie2018rethinking} extends the idea of SE-Net by squeezing along spatio-temporal axes for video processing. As the temporal relation generally plays the key role in video understanding, temporal excitation and aggregation (TEA) \cite{li2020tea} leverages the context under the temporal perspective. However, since the video content can be analysed from different perspectives, the single context utilization is incomplete for modeling diverse videos. To remedy this, there are also some attempts that explore multiple contexts. For example, gather-excite network (GE-Net) \cite{hu2018gather} generalizes SE-Net with various levels of context granularity. Convolutional block attention module (CBAM) \cite{woo2018cbam} sequentially inserts its channel and spatial attention units to the backbone. 

\begin{figure}
\centering
\includegraphics[width=0.49\textwidth]{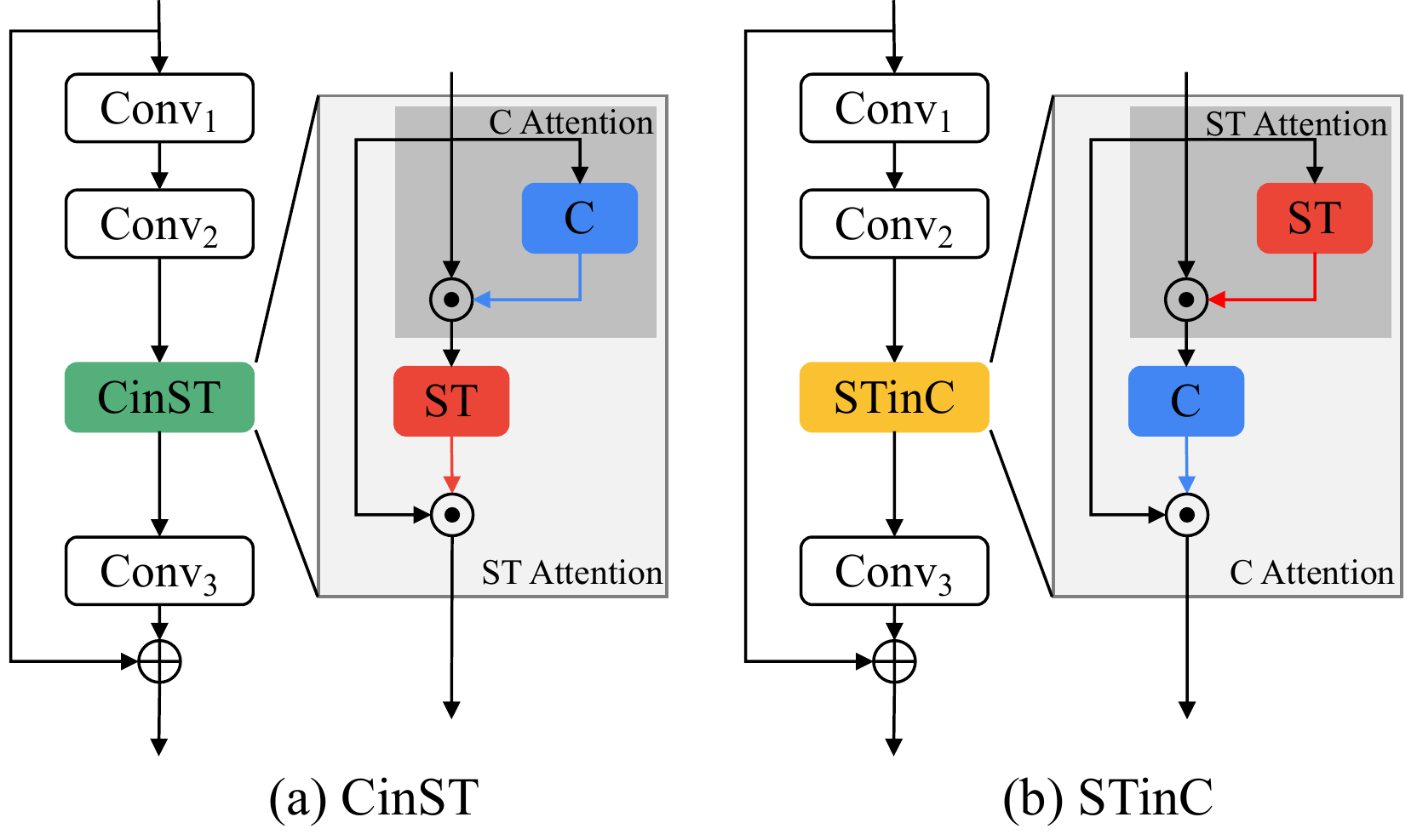}
\caption{Overall pipelines of the proposed AIA modules incorporated in the Residual block.}
\label{fig:Blocks}
\end{figure}

All the above methods share the same paradigm that different contexts work independently for generating attention map. In other words, the attention map generation is used to be all-at-once expression, which have no consideration for the correlation between different contexts. This attention fashion may be  sub-optimal since the feature axis being refined has little idea of whether the contextual information from other axes is suitable or not. This could happen when the attention result is not what the feature expects. For example, when recognizing activities that require to catch the subtle movement changes over time axis, the global information pooled over the entire spatio-temporal axes as what has been done by SE-Net will mess up those sub-activities. In this case, the under-refined channels have to accept the rough adjustment passively with nothing being done for reducing the ill impact.

To tackle this issue, we proposes Attention in Attention (AIA) modules, for explicitly and purposefully modeling the correlation between different video axial contexts. AIA mainly consists of two parts: contexts formulation and correlation capturing. Specifically, we design a family of attention units by utilizing various global axial contexts from CNN-based video features, and insert a different attention unit into an anchor attention unit for guiding the attention map learning of the anchor attention unit.

In terms of contexts formulation, we purposely derive four kinds of global contexts from a video feature map by pooling along four different axes. This pooling strategy is thus distinct from the SE-Net that compresses the entire image to one pixel. We further divide these contexts into two groups, i.e., channel group with \{channel context\} and spatio-temporal group with \{time-T, space-H, space-W contexts\} based on which two single attention units, i.e., channel attention (C) unit and spatio-temporal attention (ST) unit, are constructed respectively. In particular, the three contexts learn attention weights individually and then merge for the final attention mask in the ST unit. In contrast to many existing works, e.g., SE-Net, GE-Net and S3D, that fixedly operates on the channel dimension, the proposed C and ST units act on their pooled dimension with unified $3\times 3\times 3$ 3D convolutions. The underlying reasons lie in two aspects: (1) it can incorporate the local information in a small receptive filed ($3\times 3\times 3$) within the untouched axes to the global pooled information; and (2) the pooled dimension has a very small size (2 in AIA modules) and thus regarding it as the channel input of a 3D convolution can significantly reduce the number of added parameters.

To explicitly explore the correlation between context groups (C and ST), we further propose to insert one attention unit into the other attention unit. Accordingly, we provide two kinds of AIA modules, including C in ST attention module, referred to as CinST, and ST in C attention module, referred to as STinC. Figure \ref{fig:Blocks} shows their overall pipelines. The AIA paradigm is designed on the basis of the conjecture that different contexts should be closely correlated with each other in attention calculation. The proposed ``A-in-B'' structure provides a possible solution to verify the conjecture, which is different from and also outperforms the extended CBAM model that shares similar spirit on pooling strategy but organizes attentions in a cascaded manner. Experimental results also support it. In addition, we also present three combinations of AIA modules through sequentially and parallelly connecting them for more significant performance gains. All the AIA variants are plug-and-play modules and can be easily inserted into general networks at any layer without introducing a heavy computational burden. For compatibility, we plug AIA modules in two representative video network backbones such as TSN \cite{wang2016temporal} and TSM \cite{lin2019tsm}. 

We summarize our contributions as bellow:
\begin{itemize}
\item \textbf{Attention in Attention modules}.~We propose a new regime named attention in attention (AIA) for video feature refinement. AIA separately explores multi-axial contexts and explicitly model the correlation between contexts for more suitable attention weight learning.

\item \textbf{Efficient and easy to use}.~All attention unit variants in AIA are designed in an efficient manner, and the attention weight generation is acted on the pooled dimension with incurring little computational burden (as low as 0.02\% extra computational cost). AIA variants are plug-and-play modules and can be easily inserted into general video CNNs at any layers.

\item \textbf{Significant performance gain}.~We verify AIA modules on five commonly used benchmarks, including Something-Something V1\&V2, Diving48, EGTEA Gaze+ and EPIC-KITCHENS datasets. Experimental results show significant performance improvements for two simple video CNN backbones (i.e., TSN and TSM).
\end{itemize}

\section{Related Work}
The proposed AIA modules are targeted at refining the video features obtained from CNN-based models. As a result, we firstly review several video CNN models, including the 2D-CNN based models and 3D-CNN based models, for clearly showing the characteristics of their features. Since AIA belongs to attention regimes, we thus give brief reviews for the visual attention literatures secondly, including the most related approaches that use axial contexts and others with non-axial contexts.

\subsection{Video Networks}

{\bf 2D-CNN based video models}. The most traditional way of categorizing video activities is to directly extend the successful 2D CNNs to process 3D video signals. Here, 2D convolutions are generally used to capture static spatial information from video frames. To further model the sequential relationships among frames, two kinds of schemes are commonly used. The first one is that simply fusing the static frame features along time axis. Representative works includes \cite{Karpathy_2014_CVPR}, \cite{wang2016temporal} and \cite{zhou2018temporal}. Particularly, \cite{Karpathy_2014_CVPR} proposes to fuse information over temporal dimension through the network, temporal segment network (TSN) \cite{wang2016temporal} adopts a segmental consensus function to average per-frame prediction scores, and temporal relation networks (TRN) \cite{zhou2018temporal} merge per-frame outputs with multi-layer perceptrons. The simple temporal fusion strategies used in these works weaken their capacity in modeling dynamic relations. The other one, by contrast, utilizes the recurrent neural networks (RNNs) that inherently possess the ability of modeling temporal relations. For example, the work \cite{donahue2015long} connects up 2D-CNN and RNN and trains it in an end-to-end fashion, and \cite{huang2019spatial} turns to employ the convolutional gated recurrent unit (ConvGRU) to model spatio-temporal features.

{\bf 3D-CNN based video models}. Currently, research efforts have been made to design unified spatio-temporal computation unit. The most straightforward way is to replace all the 2D spatial convolutions of a 2D-CNN with 3D spatio-temporal convolutions. C3D \cite{tran2015learning} and I3D \cite{carreira2017quo} are two examples of this category, where I3D further inflates the pretrained 2D convolution to its corresponding 3D convolution for network initialization. V4D \cite{zhang2020v4d} even tries to use 4D convolution to additionally capture the relations among sub-clips. Though promising in spatio-temporal relation modeling, 3D/4D convolutions incur a tremendous computational burden. More recently, there are various attempts that construct efficient spatio-temporal units to tackle the above problem. Example works include P3D \cite{qiu2017learning}, R(2+1)D \cite{tran2018closer}, SlowFast \cite{feichtenhofer2018slowfast}, X3D \cite{feichtenhofer2020x3d}, TSM \cite{lin2019tsm}, RubikShift \cite{fan2020rubiksnet}, GST \cite{luo2019grouped}, GSM \cite{sudhakaran2020gate}, etc. P3D and R(2+1)D reduce the number of parameters by decomposing the 3D spatio-temporal convolution into a 2D spatial+1D temporal operation. SlowFast introduces dual-path CNNs to operate on different sampling frequencies. X3D presents an efficient strategy to search for optimal settings for space, time, width and depth. TSM utilizes the parameter-free temporal shift operation to achieve temporal modeling. GST decomposes the feature channels into spatial and spatio-temporal groups for efficient computation. GSM combines the efficient strategies of TSM and GST for more improved network architecture.

\subsection{Attention Mechanisms}

{\bf Visual attention with axial contexts}. Visual attention aims to improve representation learning with various contexts. The attention works most relevant to our AIA are axial context based. Axial contexts are referred to as information aggregated from a (multiple) visual feature axis (axes). Firstly proposed in SE-Net\cite{hu2018squeeze}, the squeeze-and-excitation mechanism, which works as a self-gating operator to element-wisely refine the features from extractors with global context, shows remarkable success in visual feature contextualization. Another advantage of the pooling-based mechanism is that it does not increase model complexity much. Sharing similar spirit with SE-Net, successors include GE-Net \cite{hu2018gather}, CBAM \cite{woo2018cbam}, triplet attention \cite{misra2021rotate},  S3D-G \cite{xie2018rethinking}, expansion-squeeze-excitation (ESE) \cite{shu2022expansion}, TEA \cite{li2020tea}, temporal adaptive module (TAM) \cite{liu2020tam} and temporal-spatial mapping \cite{song2019temporal}. Specifically, GE-Net is the generalized version of SE-Net and investigates various levels of spatial context granularity. CBAM derives two kinds of attention units to explore both channel and spatial contexts and further arranges them in a sequential manner. Triplet attention combines multiple contexts to learn the gating mask. These works are focused on processing images. To leverage video content processing, S3D-G and ESE borrow the idea of SE-Net that compresses the entire video as a single voxel. TEA and TAM also perform average pooling but only along time axis to collapse spatial information towards time axis. Temporal-spatial mapping incorporates the temporal attention into a head ConvNet and uses the max pooling operation to compute the attention weight. Although our approach has similar consideration on context exploration with these works, AIA combines different types of axial contexts in a joint unit and arranges attention units in a cascade manner making full use of various contexts.

In addition, some works adopts computational units to adaptively learn the spatio-temporal information from a small region, like two-stream collaborative learning with spatial-temporal attention (TCLSTA) \cite{peng2018two}, stagNet  \cite{qi2019stagnet} and \cite{li2019recurrent}. In particular, all the three works utilize RNN or LSTM to obtain temporal contexts. Besides, TCLSTA further uses a set of convolutions to model spatial context, and stagNet regards the spatial objects in a keyframe as an object sequence and thus models the spatial context with LSTM.

{\bf Visual attention with non-axial contexts}. There are also some other works that explore non-aggregated visual contexts. For example,  non-local network \cite{wang2018non} pairwisely compares feature points in a tensor map and recomputes each feature vector as a weighted sum of responses across all map positions. Region-based non-local (RNL) \cite{huang2021region} network extends the non-local network by using a relative larger convolutional kernel for enhancing region features. Compact bilinear augmented query structured attention (CBA-QSA) \cite{hao2020compact} uses learnable queries to attend on key spatio-temporal locations avoiding the inefficient pairwise comparing.

\section{Attention in Attention Modules}
The proposed attention in attention (AIA) modules refine the video feature map with various axial contexts. Considering the tensor shape of a video feature map $\textbf{X}$, i.e., $C\times T\times H\times W$, we can instantiate four kinds of contexts by solely squeezing along a specific axis.  Specifically, we adopt average-pooling ($AvgPool$) and max-pooling ($MaxPool$) operations to explore both mean and max statistical responses. Formally, we have 
\begin{equation}
    \textbf{G}_{C} = Concat\left (AvgPool(\textbf{X},C), MaxPool(\textbf{X},C) \right ),
\end{equation}
\begin{equation}
\label{gt}
    \textbf{G}_{T} = Concat\left (AvgPool(\textbf{X},T), MaxPool(\textbf{X},T) \right ),
\end{equation}
\begin{equation}
\label{gh}
    \textbf{G}_{H} = Concat\left (AvgPool(\textbf{X},H), MaxPool(\textbf{X},H) \right ),
\end{equation}
\begin{equation}
\label{gw}
    \textbf{G}_{W} = Concat\left (AvgPool(\textbf{X},W), MaxPool(\textbf{X},W) \right ),
\end{equation}
where $\textbf{G}_{C}\in \mathbb{R}^{2\times T\times H\times W}$, $\textbf{G}_{T}\in \mathbb{R}^{C\times 2\times H\times W}$, $\textbf{G}_{H}\in \mathbb{R}^{C\times T\times 2\times W}$ and $\textbf{G}_{W}\in \mathbb{R}^{C\times T\times H\times 2}$, and $Concat(\textbf{A}, \textbf{B})$ concatenates  $\textbf{A}$ and $\textbf{B}$ along the squeezing dimension.

The resulting contexts $\left \{ \textbf{G}_{C}, \textbf{G}_{T}, \textbf{G}_{H}, \textbf{G}_{W} \right \}$ represent the global statistics of a video feature when viewing from different perspectives. Generally, we can divide them into two context groups based on their feature attributes, i.e., channel group $\left \{ \textbf{G}_{C} \right \}$ and spatio-temporal group $\left \{ \textbf{G}_{T}, \textbf{G}_{H}, \textbf{G}_{W} \right \}$. The channel group reflects the aggregated channel information while the spatio-temporal group reflects the aggregated spatio-temporal information of video features. Based on this categorization, we can thus build two attention units: attention with channel context, referred to as \textbf{C} unit, and attention with spatio-temporal context, referred to as \textbf{ST} unit. As we analyzed in Introduction, the single context utilization unit (C or ST) ignores the correlation between different types of contexts. Based on this, we accordingly propose two attention in attention modules, i.e., CinST and STinC, where an attention unit (C/ST) is incorporated into the other attention unit (ST/C) for full utilization of different contexts. In the followings, we elaborate these single and AIA attention modules.

\subsection{Single Attention Unit, C and ST}
The single attention units C and ST separately operate on the contextual feature groups $\left \{ \textbf{G}_{C} \right \}$ and $\left \{ \textbf{G}_{T}, \textbf{G}_{H}, \textbf{G}_{W} \right \}$ with the $3\times 3\times 3$ 3D convolution. The input channel dimension of the 3D convolution is the pooling dimension, whose size is $2$, and the output channel size is set to $1$. Therefore, we can easily calculate the parameters of the used 3D convolution and that is $3\times 3\times 3 \times (C_{in}=2) \times (C_{out}=1)=54$. This is an extremely small number of parameters compared to the convolution in a regular residual block. Then, the $Sigmoid$ function is used to compute a $0-1$ gating weight based the element in the output of the 3D convolution. Next, we expand the gating weights to have the same size with the input feature $\textbf{X}$ by element copying, yielding an attention mask $\textbf{Y}$ for $\textbf{X}$. The computation pipeline is formulized as 
\begin{equation}
    \textbf{Y}_{C} = Expand\left (Sigmoid\left (  3DConv\left ( \textbf{G}_{C} \right )\right ) \right ),
\end{equation}
and
\begin{equation}
\label{yt}
    \textbf{Y}_{T} = Expand\left (Sigmoid\left (  3DConv\left ( \textbf{G}_{T} \right )\right ) \right ),
\end{equation}
\begin{equation}
\label{yh}
    \textbf{Y}_{H} = Expand\left (Sigmoid\left (  3DConv\left ( \textbf{G}_{H} \right )\right ) \right ),
\end{equation}
\begin{equation}
\label{yw}
    \textbf{Y}_{W} = Expand\left (Sigmoid\left (  3DConv\left ( \textbf{G}_{W} \right )\right ) \right ),
\end{equation}
\begin{equation}
\label{yst}
    \textbf{Y}_{ST} = \frac{1}{3}\left ( \textbf{Y}_{T} + \textbf{Y}_{H} + \textbf{Y}_{W} \right ).
\end{equation}
Finally, the refined feature $\textbf{Z}$ are obtained in an element-wise gating manner and we have
\begin{equation}
\label{zc}
    \textbf{Z}_{C} = \textbf{Y}_{C}\odot \textbf{X},
\end{equation}
\begin{equation}
\label{zst}
    \textbf{Z}_{ST} = \textbf{Y}_{ST}\odot \textbf{X},
\end{equation}
where $\odot$ denotes the element-wise multiplication.

In general, both C and ST attention units compute their 1-dimensional attention response in a $3\times 3\times 3$ receptive filed, which means that the global contexts (obtained with average-pooling and max-pooling operations) are further compressed under a local neighbourhood. Given the C attention unit as an instance, the squeezed 2-dimensional channel statistics further slims down to a 1-dimensional channel result with the 3D convolution that slides the kernel along spatio-temporal axes. As a result, besides the global aggregated information, the resulting 1-dimensional channel context additionally contains the spatio-temporal local information. It thus enables a 2D ResNet model to have the ability of spatio-temporal modeling, which is also demonstrated by the experimental result that the extreme lightweight C unit can significantly improve the video classification performance of a 2D ResNet model by a large margin of $>20\%$.

\begin{figure*}
\centering
\includegraphics[width=0.76\textwidth]{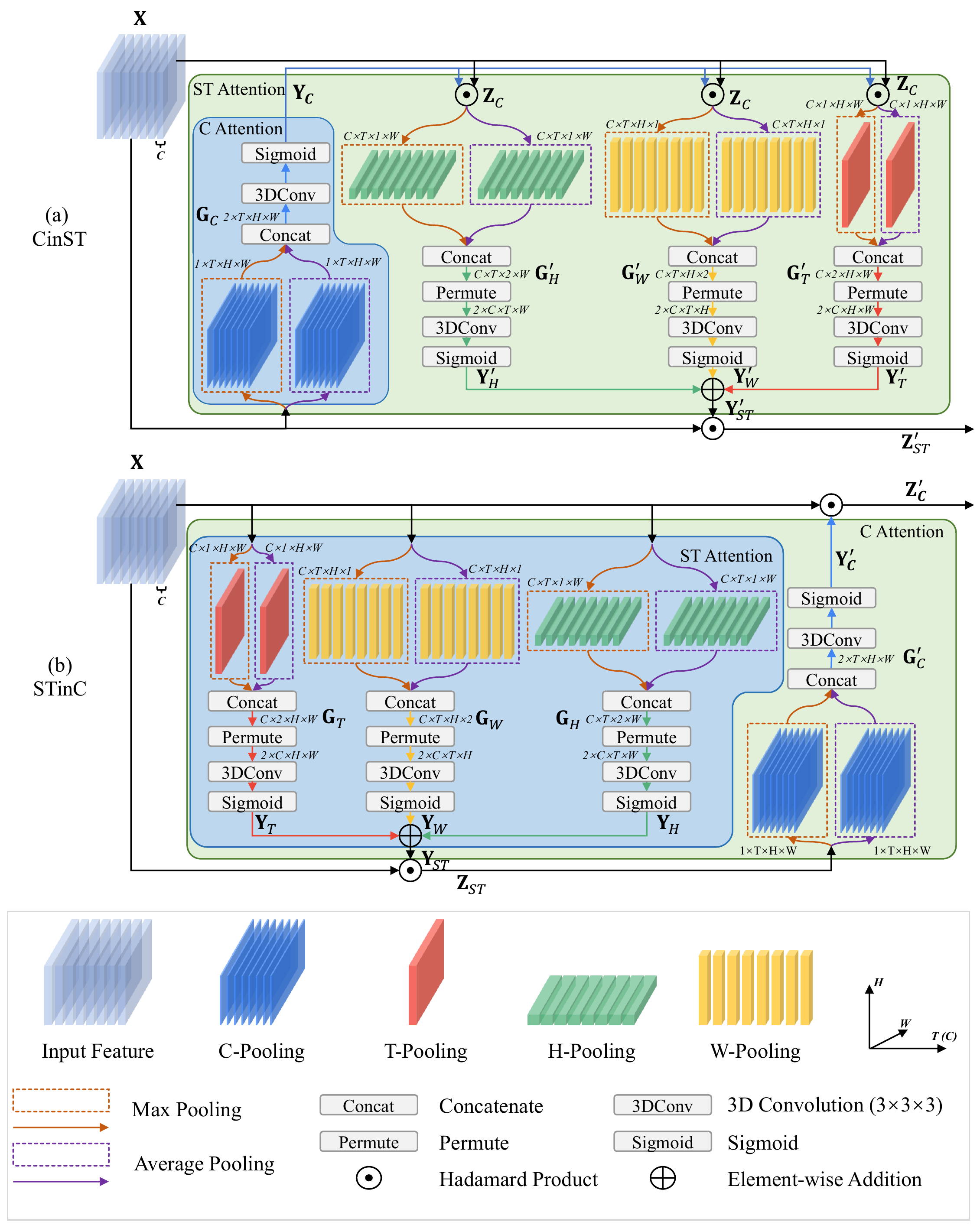}
\caption{Architectures of the proposed CinST and STinC modules.}
\label{fig:AIA}
\end{figure*}

\subsection{AIA Modules, CinST and STinC}
AIA modules take advantages of both the channel and spatio-temporal contexts to achieve video feature refinement. Different to the simple cascade and parallel connection of them as done by CBAM \cite{woo2018cbam}, we propose to insert one attention unit into the other, exploring their correlation. In the final, we also empirically combine the two AIA modules to yield further performance gains.

\subsubsection{CinST} CinST module regards the ST unit as the host attention and employs the C unit to refine the source feature $\textbf{X}$ firstly. Figure \ref{fig:AIA}-(a) illustrates the workflow details of CinST. Specifically, the obtained refined feature $\textbf{Z}_{C}$ computed from Eq. (\ref{zc}) is used as the input tensor of ST unit instead of the original feature map $\textbf{X}$ in Eqs. (\ref{gt}-\ref{gw}). Consequently, the new global contexts $\left \{ \textbf{G}'_{T}, \textbf{G}'_{W}, \textbf{G}'_{W}\right \}$ are recomputed as 
\begin{equation}
\label{gt2}
    \textbf{G}'_{*} = Concat\left (AvgPool(\textbf{Z}_{C},*), MaxPool(\textbf{Z}_{C},*) \right ),
\end{equation}
where $*$ can be $T$, $H$ or $W$. The subsequent calculations for the attention mask follow Eqs. (\ref{yt}-\ref{yst}, \ref{zst}) but with the new global contexts $\left \{ \textbf{G}'_{T}, \textbf{G}'_{W}, \textbf{G}'_{W}\right \}$ as input.

\subsubsection{STinC} In contrast to CinST, STinC takes the C unit as the host attention and accordingly adopts the ST unit to refine the source feature $\textbf{X}$, as shown in Figure \ref{fig:AIA}-(b). Here, the refined feature $\textbf{Z}_{ST}$ by Eq. (\ref{zst}) is regarded as the input tensor of C unit. The new $\textbf{G}'_{C}$, $\textbf{Y}'_{C}$ and $\textbf{Z}'_{C}$ are therefore computed as 
\begin{equation}
    \textbf{G}'_{C} = Concat\left (AvgPool(\textbf{Z}_{ST},C), MaxPool(\textbf{Z}_{ST},C) \right ),
\end{equation}
\begin{equation}
    \textbf{Y}'_{C} = Expand\left (Sigmoid\left (3DConv\left ( \textbf{G}'_{C} \right )\right ) \right ),
\end{equation}
\begin{equation}
\label{zc2}
    \textbf{Z}'_{C} = \textbf{Y}'_{C}\odot \textbf{X}.
\end{equation}

It is worth noting that all the calculations in CinST and STinC are acted on the attention weights and do not change the original feature map $\textbf{X}$ before computing the final feature map $\textbf{Z}$. In other words, the two single attention units C and ST in CinST/STinC become an organic whole. That is why we call CinST/STinC as an attention in attention module.

\begin{figure*}[b]
\centering
\includegraphics[width=0.95\textwidth]{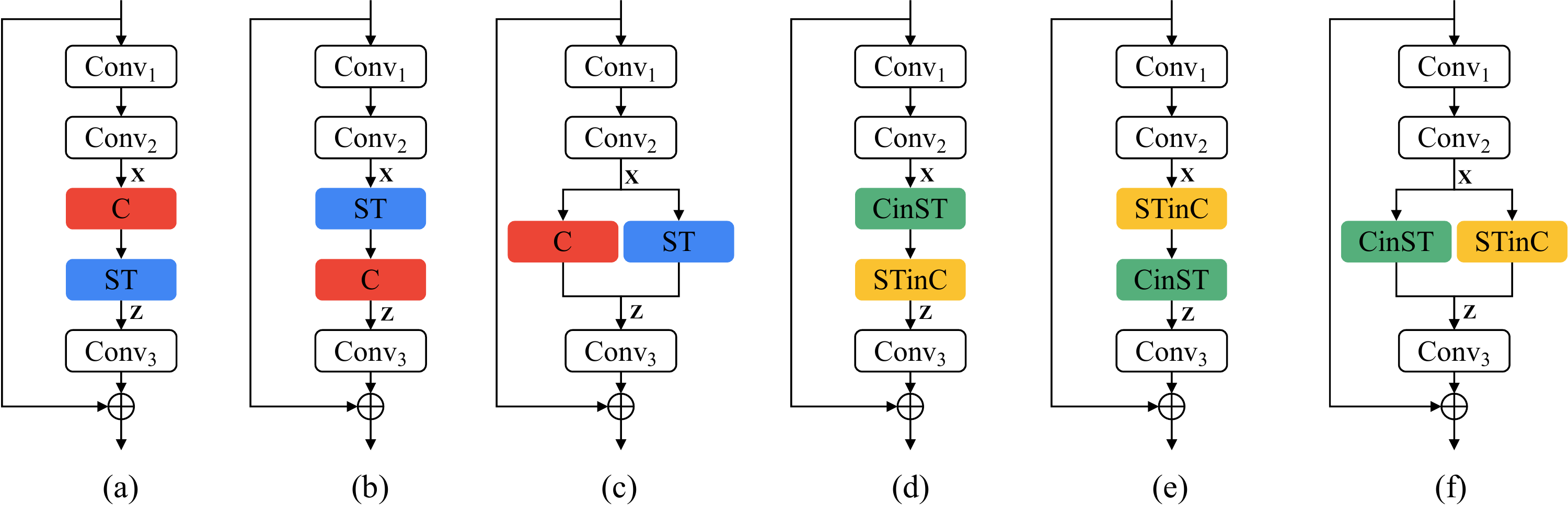}
\caption{Integration of attention modules into a ResNet block.}
\label{fig:Cascade2}
\end{figure*}

\subsection{Attention Combination and Network Architecture}
The two attention modules CinST and STinC can work either solely or cooperatively for video feature refinement. In this section, we present three combinations as shown in Figure \ref{fig:Cascade2}. All the proposed AIA variants are plug-and-play modules and can be easily integrated into existing video network architectures (e.g., TSN \cite{wang2016temporal} and TSM \cite{lin2019tsm}). Here, we explain the three combinations of AIA modules with the referenced ResNet block. In particular, the three subfigures (d), (e) and (f) in Figure \ref{fig:Cascade2} show their combination mechanisms respectively, i.e., two types of sequential connection: (d) CinST$\rightarrow$STinC and (e) STinC$\rightarrow$CinST, and one type of parallel connection: (f) CinST$+$STinC. For comparison, we also give the integration blocks with the similar connections of the two single attention units C and ST as counterparts, illustrated by Figure \ref{fig:Cascade2}-(a), (b) and (c).

\subsection{Discussion}
The most related works to our AIA modules include SE-Net \cite{hu2018squeeze}, GE-Net \cite{hu2018gather}, CBAM \cite{woo2018cbam} and S3D-G \cite{xie2018rethinking}, which use the squeeze-and-excitation scheme for visual feature refinement. Among these models, SE-Net, GE-Net and CBAM are designed for image classification tasks and S3D-G is an extension of SE-Net but works for video processing. Attention modules in SE-Net, GE-Net and S3D-G utilize parametric operators to act on the channel dimension, while CBAM and our AIA modules act on the pooled dimension requiring much lower computational burden (only tens or hundreds parameters). In addition, all the above attention modules are single attention variants, whereas our CinST and STinC are attention in attention modules. These differences are attended from the aspect of model architecture. In terms of the feature refinement mechanism, the attention modules produced by the existing works generally redistributes the global pooled information to local features (e.g., channels), while in contrast our AIA modules incorporate the local information preserved in a small receptive filed within the untouched axes to the global pooled information. In terms of motivation, our AIA modules uses the ``A-in-B'' structure to explore the correlation between two contexts, which is ignored by the other works. In the experiment, we also show the superior performances of our AIA modules to SE-Net, GE-Net, S3D-G and CBAM.

\begin{figure*}
\centering
\includegraphics[width=0.8\textwidth]{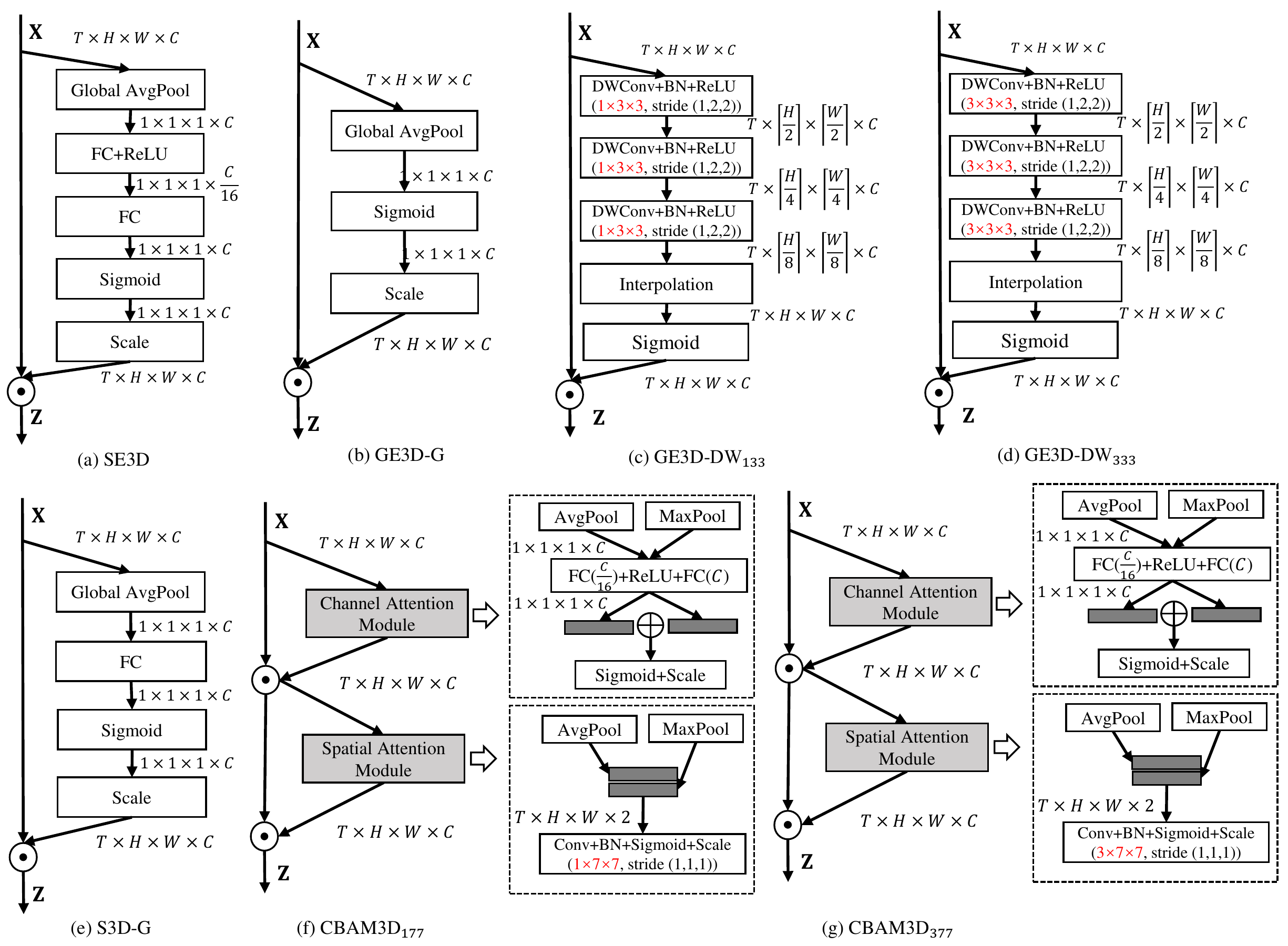}
\caption{Architectures of SE3D, GE3D-G, GE3D-DW$_{133}$, GE3D-DW$_{333}$, S3D-G, CBAM3D$_{177}$ and CBAM3D$_{377}$. ``DWConv'' denotes the depthwise convolution.}
\label{fig:seges3d}
\end{figure*}

\section{Experiments}
We conduct extensive experiments on five standard benchmarks for video classification and evaluate the performance with top-1/5 accuracy (\%). We also report the number of parameters and FLOPs to clearly show model complexity. Here, FLOPs describe how many operations are required to run a single instance of a given model and are machine-independent. Fewer FLOPs indicate that the model requires fewer computational operations and thus is more efficient.

\subsection{Datasets}
{\bf Something-Something V1 and V2}. Something-Something V1 \cite{goyal2017something} and V2 \cite{mahdisoltani2018fine} (SSV1 and SSV2 for short) datasets have the same 174 action categories and only differ in data scale. Specifically, SSV1/V2 contains $\sim$108k/220k videos, with $\sim$86k/169k in training set and $\sim$12k/25k in validating set, respectively. Videos in the datasets show fine-grained human performing actions that occur in the physical world. It requires strong temporal modeling for understanding.

{\bf Diving48}. Diving48 \cite{li2018resound} is a fine-grained video dataset of competitive diving, consisting of $\sim$18k trimmed video clips of 48 unambiguous dive sequences. The dataset is partitioned randomly into a training set of $\sim$16k videos and a test set of $\sim$2k. Dives may differ in stages (takeoff, flight, entry) and thus require modeling of long-term temporal dynamics. The used dataset is the recently cleaned version (V2).

{\bf EGTEA Gaze+}. EGTEA Gaze+ \cite{Li_2018_ECCV} dataset consists of $\sim$10k first-person vision instances for 106 non-scripted activity classes. We use the three official train/validation splits for performance report. 

{\bf EPIC-KITCHENS}. EPIC-KITCHENS \cite{damen2018scaling} also offers first-person vision actions that happen in the kitchen but focuses on object level visual reasoning in videos. In this work, we select the EPIC-KITCHENS-55 for use and separately report the verb and noun classification results. The train/validation partition is done by ourselves following the similar strategy in \cite{Baradel_2018_ECCV}.

\begin{table}
\centering
\caption{Comparisons of performance, parameters and FLOPs of different modules on Something-Something V1 dataset with 8 frames/video and the $224\times 224$ center crop.}
\begin{tabular}{c|l|c|cc|c}
\hline
Model &  Block            & Params           & Top-1   & Top-5 & FLOPS \\ \hline\hline 
\multirow{17}{*}{TSN} & None   & 23.86M     & 19.7   &  46.6     &32.88G \\  %
   &  SE3D                     & 26.38M     & 22.0    &   49.9        & 32.94G  \\  %
  &  GE3D-G                    & 23.86M     & 22.3    &   51.1        & 32.92G  \\ %
  &  GE3D-DW$_{133}$           & 24.36M     & 19.4    &   46.7     & 33.04G  \\  %
  &  GE3D-DW$_{333}$           & 25.18M     & 44.2     &   72.6      & 33.31G  \\  %
  &  S3D-G                     & 25.13M     & 28.9   &   59.3       & 32.89G  \\  %
  &  CBAM3D$_{177}$            & 26.39M     & 28.1   &   57.6     & 32.94G  \\  %
  &  CBAM3D$_{377}$            & 26.40M     & 43.4   &   72.2      & 32.96G  \\  %
  \cline{2-6}
   &  C                        & 23.87M     & 44.3    &   72.7     & 32.88G  \\ %
   &  ST                       & 23.87M     &  43.7    & 72.5      & 33.01G  \\ %
   &  C$\rightarrow$ST         & 23.87M     &  44.3   &  73.2     & 33.01G  \\  %
   &  ST$\rightarrow$C         & 23.87M     &  44.3   &   73.0      & 33.01G  \\  %
   &  C+ST                     & 23.87M     &  43.6   &  72.5      & 33.01G  \\  %
   &  CinST                    & 23.87M     & 46.9    &   75.6     & 33.01G  \\ %
   &  STinC                    & 23.87M     &  44.5   & 73.2      & 33.01G  \\  %
   &  CinST$\rightarrow$STinC  & 23.87M     & \textbf{48.5} & \textbf{77.2} & 33.15G  \\ %
   &  STinC$\rightarrow$CinST  & 23.87M     & 47.1 & 75.8 & 33.15G  \\  %
   &  CinST+STinC              & 23.87M     & 46.9 & 75.5  & 33.15G  \\  %
\hline\hline
\multirow{10}{*}{TSM}          &  None      & 23.86M     & 45.6   &  74.2   & 32.88G  \\ %
   &  C                        & 23.87M     & 47.9  &  76.3    & 32.88G  \\ %
   &  ST                       & 23.87M     &  48.1   & 76.8   & 33.01G  \\  %
   &  C$\rightarrow$ST         & 23.87M     &   48.3  &  77.1    & 33.01G \\  %
   &  ST$\rightarrow$C         & 23.87M     &  48.3    & 77.0    & 33.01G  \\  %
   &  C+ST                     & 23.87M     &  48.3    & 77.1    & 33.01G  \\  %
   &  CinST                    & 23.87M     & 48.7 & 77.4 & 33.01G  \\  %
   &  STinC                    & 23.87M     & 48.4 & \textbf{77.8} & 33.01G  \\ %
   &  CinST$\rightarrow$STinC  & 23.87M     & \textbf{49.2} & 77.5 & 33.01G  \\  %
   &  STinC$\rightarrow$CinST  & 23.87M     & 48.6 & 77.1 & 33.15G  \\  %
   &  CinST+STinC              & 23.87M     & 49.1 & 77.4 & 33.15G  \\  %
\bottomrule
\end{tabular}
\label{ab1}
\end{table}

\subsection{Implementation Details}
We implement our AIA modules on TSN and TSM backbones which are based on ResNet-50 and pretrained on ImageNet-1k dataset. We add a ``Batch-Norm'' after each convolution in AIA modules. We follow the training/inference protocols described in TSN \cite{wang2016temporal} to conduct experiments. In particular, video frames are initially resized with 240$\times$320 for SSV1/V2 and with the short-size as 256 for others before inputting into networks. During training, a $224\times 224$ center crop is used as the model input. While for testing, we infer the classification performance using center crops of a single clip in ablation study and multiple clips in the final performance report. The settings of sampled frames and clips will be specified in the tables. The training settings are set as: 50 epochs with batch size 8 per GPU, learning rate (lr) 0.01, decaying lr by 0.1 at epoch 20 and 40. All the results are obtained with Pytorch codes running on 4$\times$2080Ti or 3090 GPUs.

\subsection{Ablation Study}
The ablation study is conducted on Something-Something V1 dataset. Here, we study the performance changes of different AIA modules and also compare them with the existing methods such as SE-Net \cite{hu2018squeeze}, GE-Net \cite{hu2018gather}, S3D-G \cite{xie2018rethinking} and CBAM \cite{woo2018cbam}. Particularly, as SE-Net, GE-Net and CBAM are proposed for image processing, we thus make some minor changes for the operations in their attention blocks to facilitate video data input. We rename them as SE3D, GE3D and CBAM3D respectively. For SE-Net, we only need to replace the original 2D spatial pooling operation with 3D spatio-temporal pooling operation. While, since there are multiple network variants in GE-Net, we select two representative modules for comparison, i.e., the global average pooling one (GE-G) and the depthwise convolution one (GE-DW). Similarly, for GE-Net, the 2D global pooling in GE-G is changed to a 3D version, referred to as GE3D-G, and the 2D depthwise convolutions in GE-DW are changed to 3D counterparts, referred to as GE3D-DW, including GE3D-DW$_{133}$ with a $1\times 3\times 3$ convolution, and GE3D-DW$_{333}$ with a $3\times 3\times 3$ convolution. Finally, for CBAM, the 2D global pooling operations is changed to a 3D version and 2D convolutions are accordingly replaced with 3D convolutions in its channel and spatial attention units, resulting in two corresponding CBAM3D variants CBAM3D$_{177}$ (using a $1\times 7\times 7$ convolution) and CBAM3D$_{377}$ (using a $3\times 7\times 7$ convolution). Figure \ref{fig:seges3d} shows their detailed architectures.

Table \ref{ab1} shows their results. Firstly, the proposed single attention units C and ST significantly improve the backbones' performances (19.5\%$\rightarrow$44.3\%/43.7\% for TSN and 45.6\%$\rightarrow$47.9\%/48.1\% for TSM) but only incur $\sim$0.003M ($<$0.012\%) extra parameters and $\sim$0.006G ($<$0.02\%) extra FLOPs. Secondly, AIA modules CinST and STinC consistently outperform the single units C and ST, and also perform better than C$\rightarrow$ST, ST$\rightarrow$C and C+ST on the two backbones, which in a sense demonstrates the effectiveness of context correlation modeling. Thirdly, the combination variants CinST$\rightarrow$STinC, STinC$\rightarrow$CinST and CinST+STinC further boost the performance of backbones to new heights (i.e., 46.9\%-48.5\% for TSN and 48.7\%-49.2\% for TSM). Among the three combination variants, CinST$\rightarrow$STinC achieves the best results on the two backbones. We thus select CinST$\rightarrow$STinC as the ultimate AIA version for the performance comparison with other methods. Finally, we give the comparison with the existing plug-in modules. As shown in this table, all the proposed attention modules far outstrip SE3D, GE3D-G/DW$_{133}$, S3D-G and CBAM3D$_{177}$, which have not the function of modeling local spatio-temporal patterns, in classification accuracy. When using a 3D convolution, GE3D-DW$_{333}$ and CBAM3D$_{377}$ obtains significant performance improvements (19.4\%$\rightarrow$44.2\% and 28.1\%$\rightarrow$43.4\%), which are still worse than ours. This gives evidence that local spatio-temporal information is very important in video content understanding.

\subsection{Example Demonstration}

\begin{figure*}
\centering
\includegraphics[width=0.98\textwidth]{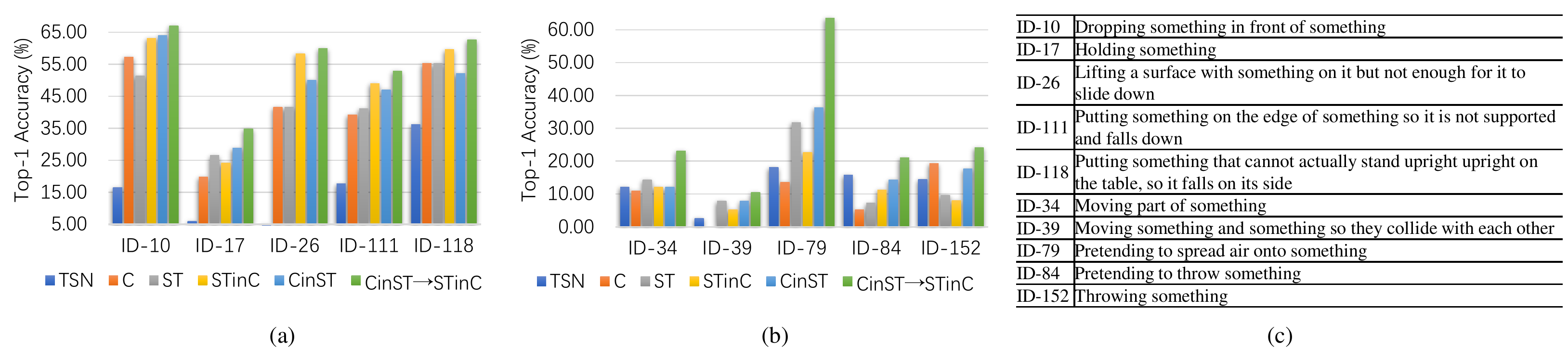}
\caption{Per-category top-1 accuracy comparison for TSN and TSN+\{C, ST, STinC, CinST, CinST$\rightarrow$STinC\} over 10 selected activity categories on Something-Something V1 dataset (validation set).}
\label{fig:select_samp}
\end{figure*}

We show the per-category results of TSN+\{C, ST, STinC, CinST, CinST$\rightarrow$STinC\} models to understand the impacts of axial contexts and the proposed AIA regime on different types of video activities in Figure \ref{fig:select_samp}. Firstly, as shown in the subfigure \ref{fig:select_samp}(a), all the attention variants significantly boost the recognition of activities that need long-range dependencies, e.g., `` \textit{ID-10: Holding something}'', ``\textit{ID-26: Lifting a surface with something on it but not enough for it to slide down}'' and ``\textit{ID-118: Putting something that cannot actually stand upright upright on the table, so it falls on its side}''. This is probably because that the global axial pooling operation can offer the attention modules the capability of modeling such long-term dependencies, which is much suitable for recognizing the continuous actions. Secondly, the proposed AIA modules STinC and CinST consistently perform better than the single attention units C and ST, which also provides evidence for the superiority of AIA. In addition, we show the failure cases that C and ST do not bring performance improvement to the backbone in the subfigure \ref{fig:select_samp}(b). Most of these activities contain short/partial motions/movements. For example, to recognize the activity of ``\textit{ID-34: Moving part of something}'', the model needs to perceive the subtle change of a small spatial region or object sensitively, however it is obviously that the C unit that smoothens the feature channels with pooling operations may harm the motion information. The other example ``\textit{ID-84: Pretending to throw something}'' only contains a very short, small amplitude at the beginning of video. Consequently, both C and ST units drop the performance. But, since we use local convolutions on the uncompressed axes, STinC and CinST, to an extent, lessen the risk of performance degradation. Finally, the the shortcomings of single attention units are made up and their advantages are mutually complemented in the cascade CinST$\rightarrow$STinC, achieving performance improvement at last.

\subsection{Visualization}

\begin{figure*}
\centering
\includegraphics[width=0.85\textwidth]{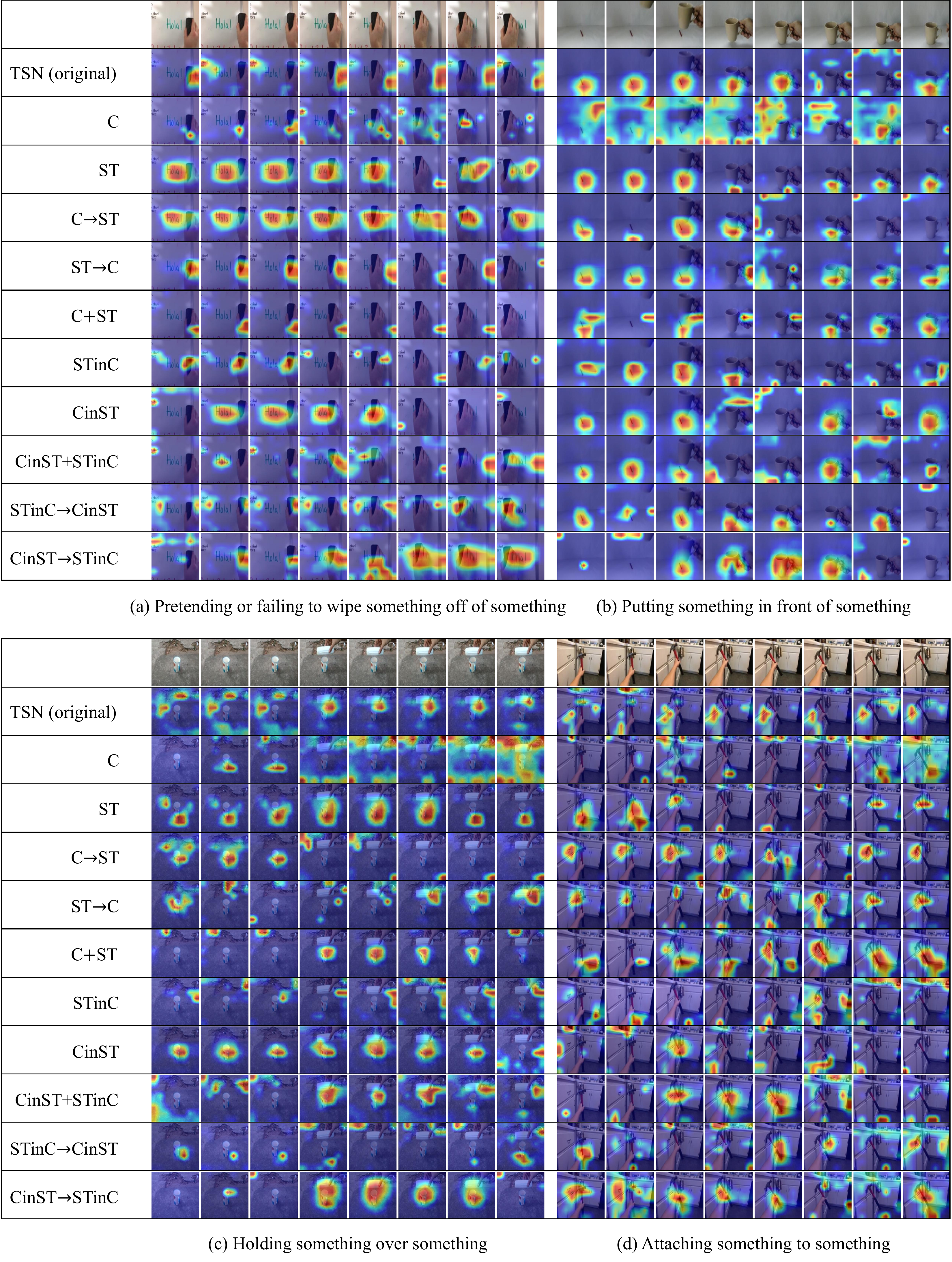}
\caption{Visualization of class activation maps on sample video clips from the Something-Something V1 dataset. The first row presents original frames and each of the other rows presents the visualization results of a model.}
\label{fig:AIA_Example}
\end{figure*}

In this section, we visualize the class activation maps using TSN with \{C, ST, C$\rightarrow$ST, ST$\rightarrow$C, C+ST, STinC, CinST, CinST+STinC, STinC$\rightarrow$CinST and CinST$\rightarrow$STinC\} to clearly show the vital parts they learn. Figure~\ref{fig:AIA_Example} shows some examples of heatmaps obtained using the Grad-CAM \cite{selvaraju2017grad} technique. Specifically, we use 8-frame center crops as input. These videos are selected from Something-Something V1 dataset. The categories in Something-Something dataset emphasize various kinds of action interactions, such as the short-term interaction between objects (e.g., ``Putting something in front of something''), the semi-intergradation (e.g., ``Holding something over something''  and ``Attaching something to something''), and the long-range interaction (``Pretending or failing to wipe something off of something''). In this case, a good model should jointly capture the interactions as completely as possible. From the visualization results, we find that the combined AIA variants, i.e., CinST+STinC, STinC$\rightarrow$CinST and CinST$\rightarrow$STinC, can mostly pick the core locations that have object interactions and CinST$\rightarrow$STinC yields more reasonable class activation maps than others in terms of these video samples.

\begin{table*}
\centering
\caption{Performance comparison of state-of-the-arts on Something V1 and V2 datasets. All the competing methods adopt ResNet-50 or its 3D/4D variants as backbone.}
\begin{tabular}{l|c|c|cc|ccccc}
\hline
\multirow{2}{*}{\textbf{Method}} & \multirow{2}{*}{\textbf{Backbone}}& \multirow{2}{*}{\textbf{Frames$\times$Crops$\times$Clips}} & \multirow{2}{*}{\textbf{Params}} & \multirow{2}{*}{\textbf{FLOPs}}& \multicolumn{2}{c}{\textbf{V1}} &  & \multicolumn{2}{c}{\textbf{V2}} \\ 
\cline{6-7} \cline{9-10} 
 &  &  &  &   & Top-1    & Top-5   &  & Top-1    & Top-5    
\\ \hline\hline
TSN~\cite{wang2016temporal}& ResNet-50 &8$\times$1$\times$1 & 23.9M &32.9G$\times$1 &19.7 &46.6 &  &30.0 &60.5\\
\hline
MFNet-C50~\cite{lee2018motion} & ResNet-50  &10$\times$1$\times$1 & ---  &  ---&40.3 &70.9 &  & ---  & ---  \\ 
\hline
I3D~\cite{carreira2017quo}  & \multirow{3}{*}{ResNet-50}&\multirow{3}{*}{32$\times$1$\times$2} & 28.0M & 153.0G$\times$2 &41.6 &72.2 &  & ---  &   ---\\
NLI3D~\cite{wang2018non}  & &  & 35.3M & 168.0G$\times$2 &44.4 &76.0 &  & ---  & ---  \\
NLI3D+GCN~\cite{wang2018videos}  & & & 62.2M & 303.0G$\times$2  &46.1 &76.8&  & ---   & ---   \\ 
\hline
RubiksNet~\cite{fan2020rubiksnet}& ResNet-50 & 8$\times$1$\times$2 & ---  & ---  &46.4&74.5&  &61.7&87.3\\ 
\hline
SlowFast \cite{feichtenhofer2019slowfast} & ResNet-50 &(4+32)$\times$3$\times$2 &32.9M  &65.7G$\times$6 & ---  & --- &  &61.9 &87.0 \\
\hline
TAM~\cite{liu2020tam} & \multirow{2}{*}{ResNet-50} &8$\times$1$\times$1 & 25.6M  & 33.0G$\times$1 &46.5 &75.8 &  & 60.5 &86.2   \\
TAM~\cite{liu2020tam} &  &16$\times$1$\times$1 & 25.6M  & 66.0G$\times$1 &47.6 &77.7 &  & 62.5  & 87.6   \\
GST~\cite{luo2019grouped} &  & 16$\times$1$\times$1 & 21.0M & 59.0G$\times$1 &48.6 &77.9 &  &62.6 &87.9\\ 
\hline
TIN~\cite{shao2020temporal} & ResNet-50 &(8+16)$\times$1$\times$1 &--- &101G$\times$1 &49.6 &78.3 &  &--- &---\\ \hline
PAN~\cite{zhang2020pan} &   &   (8+8)$\times$ 4 & ---  & 67.7 $\times$ 1 &50.5 &79.2&  &63.8 &88.6\\ \hline
ABM~\cite{zhu2019approximated} &   & 16$\times$3$\times$1 &67M & ---  &49.8& ---  && ---  & ---  \\ 
\hline
TSM~\cite{lin2019tsm} & \multirow{2}{*}{ResNet-50} & 8$\times$1$\times$2 & 23.9M & 32.9G$\times$2 &47.3 &76.2 &  &61.7 &87.4 \\
TSM~\cite{lin2019tsm} &    & 16$\times$1$\times$2 & 23.9M & 65.8G$\times$2 &48.4 &78.1 &  &63.1 &88.2 \\
TSM~\cite{lin2019tsm} &    & (8+16)$\times$1$\times$2 & 23.9M & 98.7G$\times$2 &50.3 &79.3 &  &64.3 &89.0 \\
\hline
TSM+TPN~\cite{yang2020temporal} & ResNet-50 &8$\times$1$\times$1 & 24.3M &33.0G$\times$1 &49.0 & --- &  &62.0 & ---  \\ \hline
TEINet~\cite{liu2020teinet} & \multirow{2}{*}{ResNet-50}   &8$\times$1$\times$1 & 30.4M  &33.0G$\times$1 &47.4 & ---  &  &61.3 & ---  \\
TEINet~\cite{liu2020teinet} &    &16$\times$1$\times$1 & 30.4M &66.0G$\times$1 &49.9& ---  &  &62.1 & ---  \\ 
\hline
SmallBig~\cite{li2020smallbignet} & \multirow{2}{*}{ResNet-50}&8$\times$3$\times$2 & ---  &57.0G$\times$6  &48.3 &78.1 &  &61.6 &87.7\\
SmallBig~\cite{li2020smallbignet} &   &(8+16)$\times$3$\times$2& ---  &171.0G$\times$6 &51.4 &80.7 &  &--- &--- \\
\hline
STM~\cite{jiang2019stm} & \multirow{2}{*}{ResNet-50} &8$\times$3$\times$10 & 24.0M & 33.3G$\times$30 &49.2 &79.3 &  &62.3 &88.8 \\
STM~\cite{jiang2019stm} &    &16$\times$3$\times$10 & 24.0M & 66.5G$\times$30 &50.7 &80.4 &  &64.2 &89.8\\ 
\hline
CorrNet-50~\cite{wang2020video}  &V4DResNet-50 &  32 $\times$ 10 & ---  & ---  &49.3& ---  &  & ---  & ---  \\
\hline
DFB-Net~\cite{martinez2019action} & ResNet-50  &16$\times$1$\times$1 & ---  & ---  &50.1 &79.5 & & ---  & ---  \\ 
\hline
V4D~\cite{zhang2020v4d} & ResNet-50 & 8$\times$3$\times$10 & ---  & ---  &50.4 & ---  &  & ---  & ---  \\ 
\hline
RNL TSM~\cite{huang2020region} & \multirow{3}{*}{ResNet-50} &8$\times$3$\times$2  &35.5M &41.2G$\times$6 &49.5 & 78.4  && ---  & ---  \\
RNL TSM~\cite{huang2020region} & &16$\times$3$\times$2  &35.5M &82.4G$\times$6 &51.0 & 80.3  && ---  & ---  \\
RNL TSM~\cite{huang2020region} &  &(8+16)$\times$3$\times$2  &--- &123.6G$\times$6 &52.7 &81.5  && ---  & ---  \\ \hline
TEA~\cite{li2020tea} & \multirow{2}{*}{ResNet-50}    &8$\times$3$\times$10 & 24.5M  & 35.0G$\times$30 &51.7 &80.5 &  &  --- &---   \\
TEA~\cite{li2020tea}  &   &16$\times$3$\times$10 & 24.5M  & 70.0G$\times$30 &52.3 &81.9 &  &  ---  & ---   \\ \hline

\hline
\multirow{11}{*}{Our AIA(TSN)} & \multirow{11}{*}{ResNet-50}   & 8$\times$1$\times$1  & 23.9M  & 33.1G$\times$1   & 48.5 & 77.2 & & 60.3 & 86.4  \\
                       & & 8$\times$1$\times$2  & 23.9M  & 33.1G$\times$2   & 49.5 & 78.4 & & 61.3 & 87.1  \\
                       & & 8$\times$3$\times$2  & 23.9M  & 33.1G$\times$6   & 49.8 & 78.5 & & 61.9 & 87.4  \\
                       & & 12$\times$1$\times$1 & 23.9M  & 49.7G$\times$1   & 49.0 & 76.4 & & 61.2 & 86.6  \\
                       & & 12$\times$1$\times$2 & 23.9M  & 49.7G$\times$2   & 49.9 & 78.2 & & 61.8 & 87.2  \\
                       & & 12$\times$3$\times$2 & 23.9M  & 49.7G$\times$6   & 50.3 & 78.6 & & 62.4 & 87.7  \\
                       & & 16$\times$1$\times$1 & 23.9M  & 66.3G$\times$1   & 49.2 & 77.9 & & 61.4 & 86.6  \\
                       & & 16$\times$1$\times$2 & 23.9M  & 66.3G$\times$2   & 50.0 & 78.6 & & 62.2 & 87.3  \\
                       & & 16$\times$3$\times$2 & 23.9M  & 66.3G$\times$6   & 50.5 & 79.2 & & 62.7 & 87.6  \\
                       & & (8+16)$\times$3$\times$2 & ---  & 99.4G$\times$6   & 52.9 & 81.0 & & 65.0 & 89.3  \\
                       & & (8+12+16)$\times$3$\times$2 & --- & 149.1G$\times$6  &53.8  &81.9   &&65.9   &89.8 \\
\hline 
\multirow{11}{*}{Our AIA(TSM)} & \multirow{11}{*}{ResNet-50}   & 8$\times$1$\times$1  & 23.9M  & 33.1G$\times$1   & 49.2 & 77.5 & & 61.7 & 87.2  \\
                       & & 8$\times$1$\times$2  & 23.9M  & 33.1G$\times$2   & 50.2 & 78.6 & & 62.7 & 87.8  \\
                       & & 8$\times$3$\times$2  & 23.9M  & 33.1G$\times$6   & 50.4 & 79.1 & & 63.4 & 88.2  \\
                       & & 12$\times$1$\times$1 & 23.9M  & 49.7G$\times$1   & 50.9 & 79.0 & & 62.5 & 87.4  \\
                       & & 12$\times$1$\times$2 & 23.9M  & 49.7G$\times$2   & 51.5 & 79.7 & & 63.4 & 88.1  \\
                       & & 12$\times$3$\times$2 & 23.9M  & 49.7G$\times$6   & 52.0 & 79.7 & & 64.0 & 88.4  \\
                       & & 16$\times$1$\times$1 & 23.9M  & 66.3G$\times$1   & 50.4 & 78.6 & & 63.0 & 87.8  \\
                       & & 16$\times$1$\times$2 & 23.9M  & 66.3G$\times$2   & 51.1 & 79.6 & & 64.0 & 88.7  \\
                       & & 16$\times$3$\times$2 & 23.9M  & 66.3G$\times$6   & 51.6 & 79.9 & & 64.3 & 88.9  \\
                       & & (8+16)$\times$3$\times$2 & ---  & 99.4G$\times$6   & 53.9 & 81.8 & & 66.7 & 90.4  \\
                       &  & (8+12+16)$\times$3$\times$2 & --- & 149.1G$\times$6  &\textbf{55.0}  &\textbf{82.4}   &&\textbf{67.2}   &\textbf{90.8}  \\

\hline

\end{tabular}
\label{tab:sth-V1V2}
\end{table*}

\begin{table}[htbp]
\centering
\caption{Performance comparison on the updated Diving48 dataset using the train/validation split V2.}
\tabcolsep=0.08cm
  \begin{tabular}{l|l|c|cc}
\hline  
\textbf{Method} & \textbf{Backbone} & \textbf{\#Frame} & \textbf{Top-1} & \textbf{Top-5} \\ 
\hline\hline
TSN & ResNet-50 & 8 & 72.4 & 96.8 \\
C3D & 3DResNet-50 & 8 & 73.4 & 96.0 \\
GST & ResNet-50 & 8 & 74.2 & 94.5 \\
TSM & ResNet-50 & 8 & 77.6 & \textbf{97.7} \\
\hline
AIA(TSN)  & ResNet-50 & 8 &  79.3 & 97.5 \\
AIA(TSM)  & ResNet-50 & 8 & \textbf{79.4} & 97.5 \\
\hline
\end{tabular}
\label{tab:diving}
\end{table}

\subsection{Comparison with SOTAs}
In this section, we compare our AIA module (CinST$\rightarrow$STinC) with state-of-the-art (SOTA) methods on the used five datasets respectively.

\subsubsection{Something-Something V1 and V2} Table \ref{tab:sth-V1V2} shows the performance comparisons. Firstly, similar to the observation from Table \ref{ab1}, our AIA module consistently outperforms its base networks, indicating that the proposed CinST$\rightarrow$STinC is capable of enhancing both the 2D (TSN) and 3D (TSM) deep networks. Surprisingly, AIA module improves the performance of the 2D TSN network by an absolute increase of 28.8\% (19.7\%$\rightarrow$48.5\% with 8$\times$1$\times$1) on SSV1 and 30.3\% (30.0\%$\rightarrow$60.3\% with 8$\times$1$\times$1) on SSV2. This strongly demonstrates the effectiveness of the proposed feature contextualization strategies for robust video representation learning. Secondly, AIA networks achieve comparable performances with the TEA. When further considering its lightweight in model complexity (parameters and FLOPs), AIA (23.87M/33.15G) is superior to the related attention based networks such as Non-local I3D (35.3M/168.0G), TAM (25.6M/33.0G), RNL TSM (35.5M/41.2G) and TEA (24.5M/35.0G). When using (8+16) frames $\times$ 6 clips as input, our AIA(TSM) achieves the best 53.9\% and 66.7\% top-1 accuracy on SSV1 and SSV2 datasets among the competing methods that adopt the same setting. Finally, we also ensemble AIA models with \{8, 12, 16\} frames and achieve the highest 55.0\%/67.2\% top-1 accuracy.

\subsubsection{Diving48} Since this newly released version of the dataset has been thoroughly cleaned, we re-run all the competing methods by ourselves, including TSN \cite{wang2016temporal}, C3D \cite{tran2015learning}, GST \cite{luo2019grouped} and TSM \cite{lin2019tsm}, for a fair comparison. Table \ref{tab:diving} shows the obtained results with 8 frames as input. Dives are the synthesis of continuous motion changes at different stages. Consequently, the simple frame-level fusion as done by TSN can achieve relatively satisfactory result. It is sure that further capturing the temporal relations among subtle body poses can further improve the recognition, which is demonstrated by the performance gain (72.4\%$\rightarrow$77.6\%) of TSM. Moreover, our AIA additionally considers the long-range context and thus achieve the best performances 79.3\% for AIA(TSN) and 79.4\% for AIA(TSM).

\begin{table}[htbp]
\centering
\caption{Performance comparison on EGTEA Gaze+ dataset using train/validation split 1/2/3. The results of I3D-2stream, R34-2stream and SAP are cited from \cite{Li_2018_ECCV}, \cite{sudhakaran2018attention} and \cite{wang2020symbiotic}, respectively. Except R34-2stream using ResNet-34 as backbone, all the other models adopt ResNet-50 as backbone.}
\begin{tabular}{l|c|ccc}
\hline
\textbf{Method} & \textbf{\#Frame} & \textbf{Split1} & \textbf{Split2} & \textbf{Split3}\\ 
\hline\hline
I3D-2stream   & 24 &55.8 &53.1 &53.6  \\
R34-2stream   & 25& 62.2 &61.5 &58.6 \\ 
SAP  &  64 & 64.1  &62.1 &62.0 \\
\hline
TSN (our impl.)  & 8 & 61.6 &58.5 &55.2  \\
C3D (our impl.)  & 8 & 62.1 &59.2 &57.0  \\
GST (our impl.) & 8 & 63.3 &61.2 &59.2 \\
TSM (our impl.)  & 8 & 63.5 &62.8 &59.5  \\
\hline
AIA(TSN) & 8 & 63.7  &62.1  & 61.5 \\
AIA(TSM) & 8 &  \textbf{64.7} &\textbf{63.3}  &\textbf{62.2}  \\
\hline
\end{tabular}
\label{tab:egtea}
\end{table}

\begin{table}[htbp]
\centering
\caption{Performance comparison on EPIC-KITCHENS-55 dataset. The results of all the methods are obtained using our train/validation split.}
\begin{tabular}{l|c|cc}
\hline
\textbf{Method} & \textbf{\#Frame} & \textbf{Verb} & \textbf{Noun} \\ 
\hline\hline
TSN (our impl.)   & 8 &  37.4   & 23.1   \\
C3D (our impl.)   &8 &  45.2   & 21.5  \\
GST (our impl.)  &8 &  46.4   & 21.1  \\
TSM (our impl.)  &8 &  48.2   & 22.9  \\
\hline
AIA(TSN)  & 8 & 49.5& 23.5 \\
AIA(TSM)  & 8 &\textbf{50.2} & \textbf{24.3} \\
\hline
\end{tabular}
\label{tab:epickit}
\end{table}

\subsubsection{EGTEA Gaze+ and EPIC-KITCHENS} Activities in the two first-person vision datasets contain rich human-object interactions occurring in native
environments. It generally requires modeling both spatial and temporal patterns to achieve activity recognition. As shown in Table \ref{tab:egtea},  when using the same number of input frames (i.e., 8), the spatial-temporal methods, such as C3D, TSM, GST and our AIAs, obtain better performance than the spatial-only TSN. Also, equipped with spatio-temporal context, our AIA models perform best among all the competing methods. On EPIC-KITCHENS, models are required to separately recognize the motion ingredient (i.e., verb) and the object ingredient (i.e., noun) of activities. Table \ref{tab:epickit} shows the performance comparison on the two terms. We observe different performance trends on verb and noun terms. That is, the 3D models C3D, GST and TSM perform significantly better than the 2D model TSN for verb recognition, while TSN obtains a better performance for noun recognition. This may be because objects need more spatial modeling rather than temporal modeling. Among all the competing methods, our AIA models achieve the best results in both verb and noun recognition tasks, which further proves its good performance on recognizing diverse activities.

\section{Conclusion}
We have presented the attention in attention (AIA) mechanism for video feature contextualization. We firstly explore various kinds of global contexts aggregated along different axes of the video feature map. Then, we construct two types of single attention units C and ST to separately operate on the grouped contextual features. To further achieve local information modeling, C and ST adopt 3D spatio-temporal convolutions to influence global contexts. Since these units regard the squeezed dimension as the channel input of 3D convolution, both C and ST are much lightweight ($<0.012\%/0.02\%$ extra parameters/FLOPs to the backbone). We also propose the idea of inserting one attention unit to the other unit to utilize the correlation information between them and empirically combine those AIA variants. In the experiment, we densely plug AIA modules into TSN and TSM backbones. Classification results on five video benchmarks show that the proposed AIA modules can significantly improve the performances of backbones by large margins (\emph{e.g.}, 28.8\%/30.3\% for TSN on SSV1/SSV2) and also perform much better than the existing attention modules (\emph{e.g.}, SE-Net, GE-Net and S3D-G). 

\section*{Acknowledgements}
The work was supported in part by the National Key Research and Development Program of China under Grant 2020YFB1406703, and by the National Natural Science Foundation of China (Grants No. 62101524 and No. U21B2026).


 

\bibliographystyle{IEEEtran}
\bibliography{aaai22}

\end{document}